\documentclass{article}

\usepackage{microtype}
\usepackage{graphicx}
\usepackage{subfigure}
\usepackage{booktabs} 

\usepackage[accepted]{icml2024}

\usepackage{amsmath}
\usepackage{amssymb}
\usepackage{mathtools}
\usepackage{amsthm}

\usepackage{url}
\usepackage{amsmath}
\usepackage{amssymb}
\usepackage{booktabs}

\usepackage{xcolor} 
\usepackage{times}
\usepackage{epsfig}
\usepackage{graphicx}
\usepackage{amsmath}
\usepackage{amssymb}
\usepackage[bb=dsserif]{mathalpha}
\usepackage{bm}

\usepackage{booktabs,colortbl,tabularx}
\usepackage{pifont}%

\usepackage{mathtools}
\usepackage{commath}
\usepackage{algorithm}
\usepackage[noend]{algpseudocode}

\usepackage{wrapfig}
\usepackage{multirow}
\usepackage{comment} 
\usepackage{enumitem}

\definecolor{citecolor}{HTML}{2980b9}
\definecolor{linkcolor}{HTML}{c0392b}

\usepackage[pagebackref=true,breaklinks=true,colorlinks,bookmarks=false,citecolor=citecolor,linkcolor=linkcolor]{hyperref}

\definecolor{Gray}{gray}{0.9}

\renewcommand{\paragraph}[1]{\noindent {\bf #1}}

\newcommand{\cmark}{\ding{51}}%
\newcommand{\xmark}{\ding{55}}%

\usepackage{xspace}
\newcommand{\methodfull}{{Long-term Spatial Prompt Tuning}\xspace}
\newcommand{\method}{{LSPT}\xspace}

\usepackage[capitalize,noabbrev]{cleveref}

\theoremstyle{plain}

\theoremstyle{definition}

\theoremstyle{remark}

\icmltitlerunning{LSPT: Long-term Spatial Prompt Tuning for Visual Representation Learning}

\begin{document}

\twocolumn[
\icmltitle{LSPT: Long-term Spatial Prompt Tuning for Visual Representation Learning}




\begin{icmlauthorlist}
\icmlauthor{Shentong Mo}{cmu,mbz}
\icmlauthor{Yansen Wang}{msr}
\icmlauthor{Xufang Luo}{msr}
\icmlauthor{Dongsheng Li}{msr}
\end{icmlauthorlist}

\icmlaffiliation{cmu}{Carnegie Mellon University}
\icmlaffiliation{mbz}{Mohamed bin Zayed University of Artificial Intelligence}
\icmlaffiliation{msr}{Microsoft Research}

\icmlcorrespondingauthor{Shentong Mo}{shentongmo@gmail.com}

\icmlkeywords{Machine Learning, ICML}

\vskip 0.3in
]



\printAffiliationsAndNotice{}  

\begin{abstract}
Visual Prompt Tuning (VPT) techniques have gained prominence for their capacity to adapt pre-trained Vision Transformers (ViTs) to downstream visual tasks using specialized learnable tokens termed as prompts. Contemporary VPT methodologies, especially when employed with self-supervised vision transformers, often default to the introduction of new learnable prompts or gated prompt tokens predominantly sourced from the model's previous block. 
A pivotal oversight in such approaches is their failure to harness the potential of long-range previous blocks as sources of prompts within each self-supervised ViT.
To bridge this crucial gap, we introduce \methodfull (\method) – a revolutionary approach to visual representation learning. 
Drawing inspiration from the intricacies of the human brain, \method ingeniously incorporates long-term gated prompts. 
This feature serves as temporal coding, curbing the risk of forgetting parameters acquired from earlier blocks. 
Further enhancing its prowess, \method brings into play patch tokens, serving as spatial coding. 
This is strategically designed to perpetually amass global features, thereby fortifying the model's prowess in distinguishing and identifying visual categories.
To validate the efficacy of our proposed method, we engaged in rigorous experimentation across 5 FGVC and 19 VTAB-1K benchmarks. Our empirical findings underscore the superiority of \method, showcasing its ability to set new benchmarks in visual prompt tuning performance.

\end{abstract}

\section{Introduction}

The rise of the Transformer architecture~\citep{vaswani2017attention} has cemented its position as the foundational module for vision-related tasks. Within this paradigm, Vision Transformers (ViTs)~\citep{dosovitskiy2021an,touvron2020deit,liu2021Swin,yuan2021tokens} have manifested remarkable dominance over traditional Convolutional Neural Networks (CNNs) across various tasks, such as image classification, object detection, and semantic segmentation. Concurrently, the success of self-supervised learning frameworks~\citep{chen2020simple,chen2021simsiam,he2019moco,grill2020bootstrap}, especially in harnessing vast reservoirs of unlabeled data, has been undeniable. Merging these two powerhouses seems instinctual, and early forays into this combination~\citep{chen2021mocov3,xie2021self-supervised,caron2021emerging} indeed show promise, despite challenges in seamless integration.

\begin{figure*}
\centering
\includegraphics[width=0.75\linewidth]{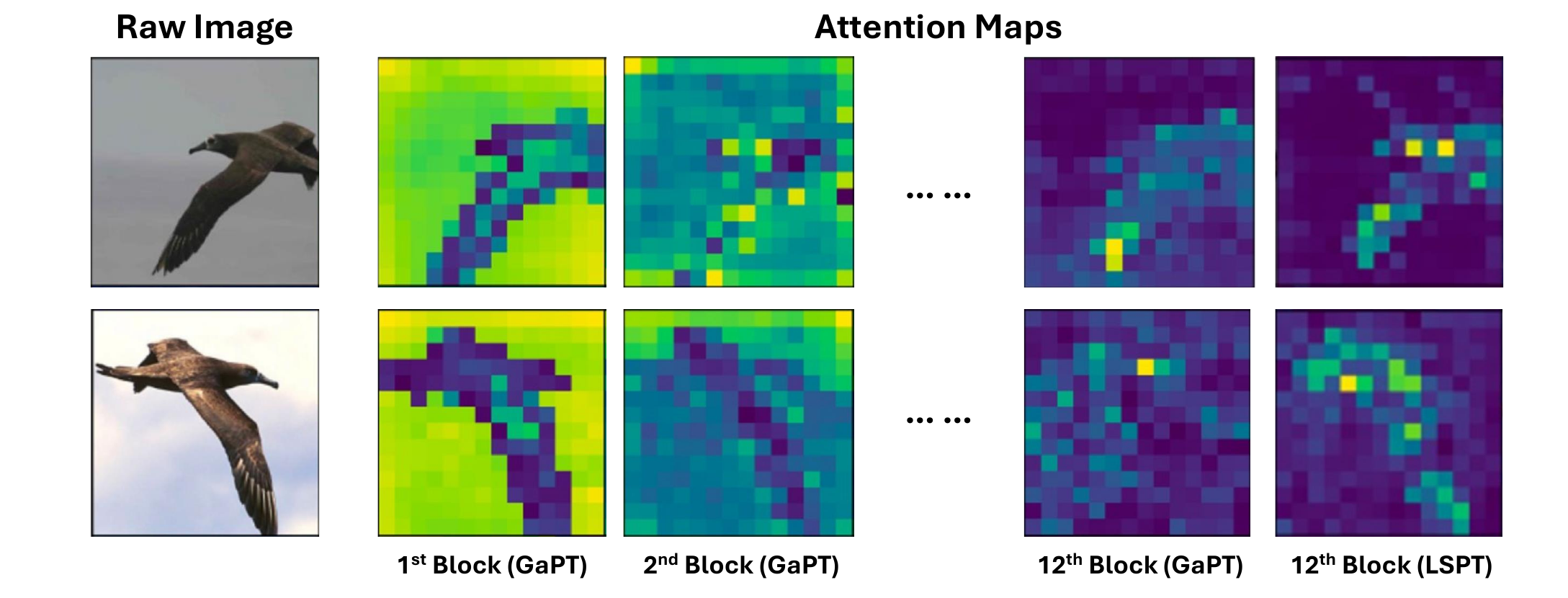}
\vspace{-1.0em}
\caption{{\bf Comparison of the forgetting problem in GaPT and the shape information awareness in our \method.} For the 12th block, the attention map of the state-of-the-art approach has been blur and almost lose the crucial spatial information. While for our \method, we can see a clear attention map for the object in the raw image, demonstrating its ability to incorporate spatial information and pass it through long-range blocks.}
\vspace{-0.3cm}
\label{fig: title_img}
\end{figure*}

Amidst this backdrop, Visual Prompt Tuning (VPT) has emerged as an influential player, adept at tailoring pre-trained ViTs for specific downstream tasks using adaptable tokens or ``prompts''. As a testament to VPT's prowess, VPT techniques prepend learnable prompts to input sequences, effectively guiding the fixed pre-trained encoder's information for task-specific objectives~\citep{jia2022vpt}. 
The Gated Prompt Tuning (GaPT) 
strategy takes this a notch further by incorporating gating mechanisms for each ViT block to modulate its influence based on cues from preceding blocks~\citep{yoo2023improving}. 

Nevertheless, some glaring limitations remain. Firstly, these methodologies largely overlook the latent potential of long-range blocks as prompt sources among blocks and cause the \textit{temporal} forgetting problem
, which also corresponds to ``long-term" forgetting across transformer blocks.
Even with the gating mechanism introduced in the gated prompt tuning strategy, the information from early blocks diminishes exponentially, making it challenging to capture in the later blocks. Additionally, the embedding of patch tokens, which encapsulates crucial spatial information and acts as an intermediate global visual representation of the image, is regrettably lost across blocks. This results in what can be termed as a \textit{spatial} forgetting phenomenon.
Figure~\ref{fig: title_img} illustrates the nature of the forgetting issue inherent in contemporary visual prompt tuning methods.
Intriguingly, these two feature forgetting challenges align remarkably with the human visual system. In humans, both temporal and spatial correlations of neuronal discharges are essential for integrating distributed neuronal activities into cohesive representations~\citep{huxter2003independent,victor1996nature,engel1992temporal,reinagel2000temporal}. Therefore, we posit that integrating both temporal and spatial coding could significantly enhance the efficacy of visual prompts.

To address these pressing concerns, we unveil the \methodfull (\method) 
framework that can explicitly alleviate the forgetting issues on both temporal and spatial aspects. 
Rooted deeply in neural mechanisms found in the human brain, \method offers a fresh perspective to visual representation learning. At its core, \method integrates long-term gated prompts, introducing a temporal coding layer that actively mitigates the forgetting of parameters learned from anterior blocks. 
By weaving in patch tokens as spatial coding elements, it additionally ensures a sustained aggregation of global features, bolstering the model's discriminative capabilities.
Subjecting \method to meticulous evaluations on 5 FGVC and 19 VTAB-1K benchmarks, we unearth empirical evidence attesting to its unparalleled prowess, setting novel standards in visual prompt tuning.

In a nutshell, our seminal contributions are:
\begin{itemize}
    \item The inception of \method: a pioneering prompt tuning paradigm adept at seamlessly integrating long-term gated prompts for temporal coding, effectively addressing the `forgetting' challenges.
    \item The novel integration of learnable spatial gated prompts, meticulously crafted to ensure a continuous accumulation of class-distinctive features.
    \item Comprehensive experimental validations that unequivocally establish \method's supremacy over existing baselines in the realm of visual prompt tuning.
\end{itemize}

\section{Related Work}

\paragraph{Self-supervised Vision Transformers}
~\citep{chen2021mocov3,xie2021self-supervised,caron2021emerging} have addressed people's attention due to their strong performance on various downstream tasks. 
Specifically,
MoCov3~\citep{chen2021mocov3} extended the MoCo~\citep{he2019moco} method to ViT~\citep{dosovitskiy2021an} for minimizing the distance between representations of two augmented views.
MoCo v2 and BYOL were applied simultaneously in MOBY~\citep{xie2021self-supervised} to form a self-supervised framework based on the Swin~\citep{liu2021Swin} backbone.
In DINO~\citep{caron2021emerging}, knowledge distillation was combined with momentum encoder and multi-crop training for learning the local-to-global correspondence in the vision transformer.
As proven to be effective in a previous study~\citep{maithra2021do}, vision transformers can obtain global representations from shallow layers.
Therefore, it is desirable to take into account low-level features from the shallow stage for learning more fine-grained invariances. 
Masked image modeling (MIM) also has been explored in many self-supervised ViTs~\citep{bao2021beit,atito2021sit,he2021masked,wei2022masked,xie2022SimMIM} to reconstruct the masked image patch given the unmasked counterpart as clues. 
For example, block-wise masking was introduced in BEiT~\citep{bao2021beit} to learn transferrable visual representations by recovering discrete tokens of masked image patches.
Given features extracted from the 25\% unmasked patches, the seminal work, MAE~\citep{he2021masked} directly reconstructed missing pixels of 75\% masked patches.
In this work, our main focus is to adapt self-supervised pre-trained vision transformers to downstream visual tasks using specialized learnable prompts, which is more challenging than fine-tuning all parameters of the pre-trained backbone architecture.

\paragraph{Visual Transfer Learning} aims to learn transferable representations from pre-trained vision backbones for downstream tasks.
Early works~\cite{dosovitskiy2021an} leveraged full fine-tuning to train both the pre-trained model and the task-specific head.
Recently, diverse parameter-efficient tuning methods have been proposed.
For example, Sidetune~\citep{zhang2020sidetuning} utilized a ``side'' network and linearly interpolated between pre-trained features and side-tuned features before being fed into a classification head.
Bias tuning~\citep{cai2020tinytl,ben2022bitfit} proposed to fine-tune only the bias terms of the pre-trained backbone.
Adapter-based approaches~\citep{houlsby2019parameterefficient,pfeiffer2020adapterfusion,pfeiffer2020adapterhub} inserted multiple MLP modules with residual connection inside visual transformer layers.
However, since these mainly deal with supervised pre-trained ViTs, few studies explored parameter-efficient tuning for self-supervised models. 
In this work, we develop a new prompt-based transfer learning method based on learnable input prompts for self-supervised ViTs.

\paragraph{Visual Prompt Tuning}
(VPT)~\citep{jia2022vpt} prepended learnable prompt tokens to the input sequences, which then act as task-specific instructions by steering the information from the fixed pre-trained encoder. VPT, when used with supervised ViT backbones, has shown outstanding performance on numerous downstream tasks.
GaPT~\citep{yoo2023improving} proposed to adapt a gate for each ViT block to adjust its intervention into the prompt tokens predominantly sourced from the model’s previous block.
However, a pivotal oversight in those approaches is their failure to harness the potential of long-range previous blocks as sources of prompts within each self-supervised ViT.
In contrast, we develop a fully novel framework to mitigate the forgetting of previously learned prompts from history transformer blocks with explicit long-term prompts and global spatial prompt coding. 
To the best of our knowledge, we are the first to leverage an explicit temporal and spatial prompts coding mechanism for visual prompt tuning. 
Our experiments in Section~\ref{sec:exp_sota} also validate the superiority of our \method in all benchmarks for prompt tuning.

\section{Preliminaries}

Given a set of downstream images, our goal is to adapt pre-trained Vision Transformers (ViTs) to downstream tasks using learnable prompt tokens. We now introduce the notations in this paper, and then revisit the VPT method.

\subsection{Notations}

A ViT generally consists of a patch embedding layer, a stack of $L$ transformer blocks, and a
classification head. 
For an input image $\mathbf{I}$ with shape of $H\times W\times 3$,
we denote the input patch tokens for the $l$-th block as $\mathbf{X}^{l-1} = [\mathbf{X}^{l-1}_1, ..., \mathbf{X}^{l-1}_N]\in\mathbb{R}^{N\times D}$, where $N = HW/P^2, l = 1,...,L$, $P$ is the patch size, and
$D$ is the dimension of the transformer blocks. $\mathbf{X}^0_i=\text{embed}(\mathbf{x}_i), i\in\{1, 2, ..., N\}$ is obtained by embedding the $i$-th patch $\mathbf{x}_i$ of the input image $\mathbf{I}$.
An additional learnable classification token $\mathbf{x}_C^{l-1}\in\mathbb{R}^{1\times D}$ is also concatenated to patch tokens for each self-attention block, that is, 
$[\mathbf{x}_C^{l},\mathbf{X}^{l}] = \mbox{AttnBlock}^{l}([\mathbf{x}_C^{l-1},\mathbf{X}^{l-1}])$, where $\mathbf{x}_C^{l}\in\mathbb{R}^{1\times D}, \mathbf{X}^{l}\in\mathbb{R}^{N\times D}$.

\begin{figure*}[t]
\centering
\includegraphics[width=0.9\linewidth]{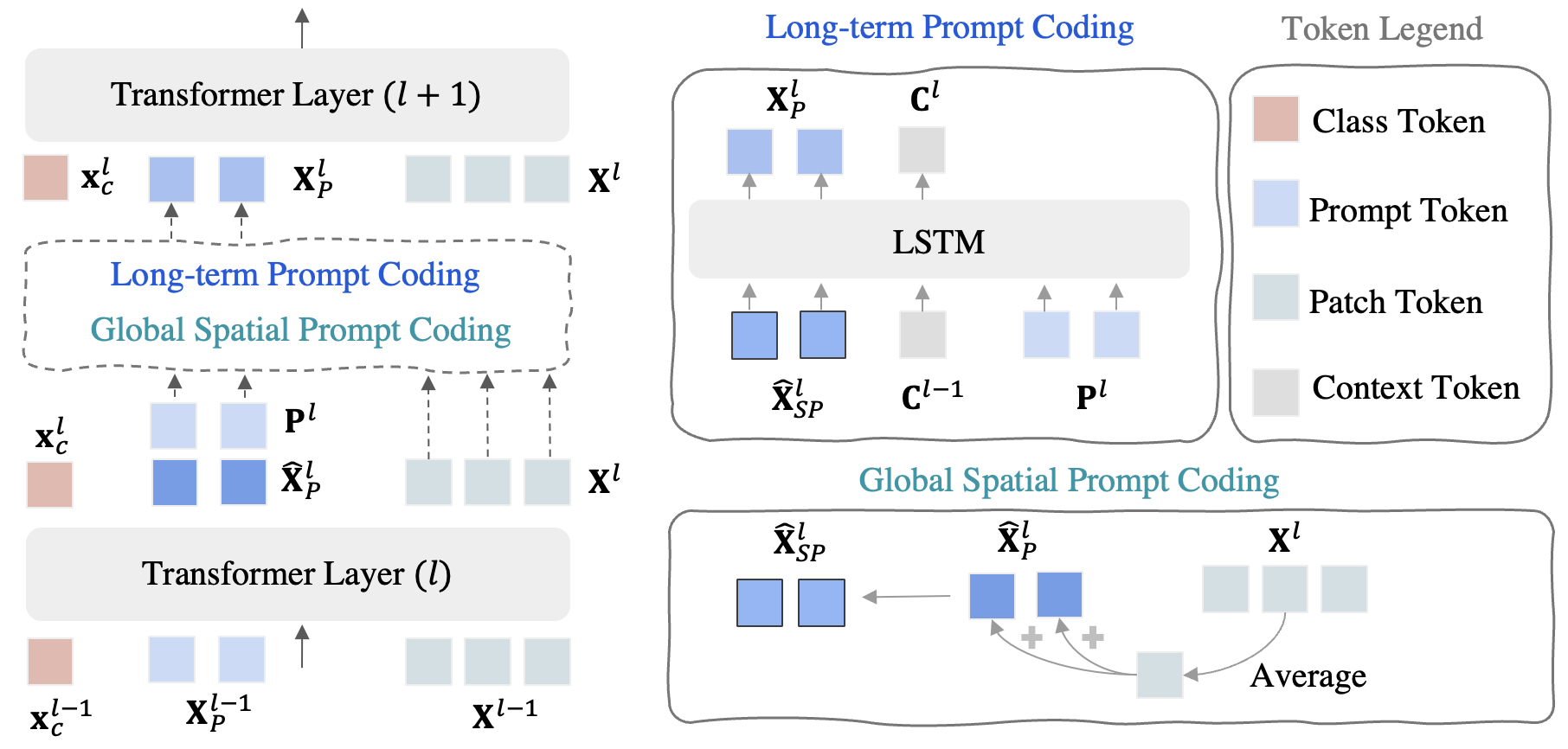}
\vspace{-1em}
\caption{{\bf Illustration of the proposed Long-term Spatial Prompt Tuning (LSPT) framework.} 
For transformer block $l$, the Global Spatial Prompt Coding (GSPC) module adds the average embeddings of patch tokens $\mathbf{X}^{l}\in\mathbb{R}^{N\times D}$ from the block to the output prompts $\widehat{\mathbf{X}}_P^{l}\in\mathbb{R}^{N_p\times D}$ to generate global spatial prompts $\widehat{\mathbf{X}}_{SP}^{l}$.
With the inserted prompt tokens $\mathbf{P}^l\in\mathbb{R}^{N_p\times D}$ and $\widehat{\mathbf{X}}_{SP}^{l}\in\mathbb{R}^{N_p\times D}$, the Long-term Prompt Coding (LPC) module with parallel importance takes the inserted prompts $\mathbf{P}^{l}$ as input and $\widehat{\mathbf{X}}_{SP}^{l}$ as hidden states, and the output context embeddings $\mathbf{C}^{l-1}\in\mathbb{R}^{N_p\times D}$ at block $l-1$ are fed into the layer as cell states.
Finally, the output updated prompts $\mathbf{X}_P^{l}$ is used as the new prompt tokens for block $l+1$ to achieve long-term prompt coding.}
\vspace{-0.5em}
\label{fig: main_img}
\end{figure*}

\subsection{Revisit Visual Prompt Tuning}

To solve the prompt tuning problem for visual classification, VPT~\citep{jia2022vpt} proposed to fine-tune continuous prompt tokens directly in the representation space and prepended these prompt tokens $\mathbf{P} = [\mathbf{p}_1,...,\mathbf{p}_{N_p}]\in\mathbb{R}^{N_p\times D}$ to the input patch tokens, where $N_p$ is the number of learnable prompt tokens and $D$ is the dimension size of the prompt tokens shared with patch tokens. 
Specifically, they froze the pre-trained ViT weights and fine-tuned the newly inserted prompt tokens $\mathbf{P}$ and a classification head for specific downstream tasks.
In order to incorporate more tuning prompts in the embedding space, VPT-deep \footnote{We illustrate VPT-deep and omit VPT-shallow here because VPT-deep is more general and performs better.} tried to inject new block-specific prompt tokens $\mathbf{P}^{l-1}=[\mathbf{p}_1^{l-1},...,\mathbf{p}_{N_p}^{l-1}]\in\mathbb{R}^{N_p\times D}$ to each block, which is formulated as
\begin{align}
[\mathbf{x}_C^{l},\mathbf{X}_P^{l},\mathbf{X}^{l}] & = \mbox{AttnBlock}^{l}([\mathbf{x}_C^{l-1},\mathbf{P}^{l-1},\mathbf{X}^{l-1}]),\quad  \label{eq:VPT-deep}
\end{align}
where $l=1,...,L$ is the block layer index. Note that $\mathbf{X}_P^{l}$ will be discarded after each block, and will not be used as input to the next block.
When training future blocks based on the output from current blocks, aggregated prompt tokens from the previous blocks might not be seen anymore.

\paragraph{Limitations of VPT}
Such a visual prompt tuning method targets learning task information for transferring. However, continually aggregating the newly injected prompt tokens in VPT is not effective enough for ViTs due to the following two aspects.
First is the \textit{temporal} forgetting problem. The global visual representation extracted from the image is catastrophically forgotten by the stack of new prompt tokens, and thus they can not associate the latest prompts with the corresponding objects in the image for future blocks.
Second is the \textit{spatial} forgetting phenomenon. They ignored the explicit incorporation of patch token embeddings and learnable prompt tokens during training, causing worse attention maps from the last transformer layer.

\section{Method}

To tackle above challenges, we argue that the injected tokens into ViTs should be able to capture both spatial information across patch tokens and temporal information among long-range blocks. Therefore, we propose a novel prompt tuning framework, namely \method, to achieve efficient tuning within self-supervised ViTs, as illustrated in Figure~\ref{fig: main_img}. 

By reviewing the design of VPT-deep in Equation \ref{eq:VPT-deep}, we notice that meaningful sources for constructing the injected prompt input into ViTs are 3-fold: 1) the newly added prompt token $\mathbf{P}^{l-1}$; 2) the output prompt tokens which contain information from anterior blocks; 3) the output patch tokens which contain spatial information of the image tokens. The first source provides learnable parameters to enhance model's expressiveness during transferring and the last two offer valuable information which can help address the temporal and spatial forgetting problem.

\method constructs the injected prompt input into ViTs (denoted by $\mathbf{X}_P^{l-1}$ for the $l$-th block) based on added prompt tokens and enhance them via two modules that explicitly utilize last two sources, respectively. Specifically, Global Spatial Prompt Coding in Section~\ref{sec:cspc} is designed for using output patch tokens to better capture spatial information. Long-term Prompt Coding in Section~\ref{sec:lpc} is for using output prompt tokens to address the temporal forgetting problem.

\subsection{Global Spatial Prompt Coding}\label{sec:cspc}

We first introduce a novel and explicit global spatial prompt coding module to incorporate global visual representation extracted from the image learned from previous self-attention blocks.
To make the prompt tokens be aware of the global information in patch embeddings $\mathbf{X}^{l-1}\in\mathbb{R}^{N\times D}$ from $l-1$th transformer block, we add the average embedding of patch tokens to all output prompts $\widehat{\mathbf{X}}_P^{l-1}\in\mathbb{R}^{N_p\times D}$ for spatial prompt coding, which is formulated as
\begin{align}
[\mathbf{x}_C^{l},\widehat{\mathbf{X}}_P^{l},\mathbf{X}^{l}] & = \mbox{AttnBlock}^{l}([\mathbf{x}_C^{l-1},\mathbf{X}_P^{l-1},\mathbf{X}^{l-1}]),  \\
\widehat{\mathbf{X}}_{SP,k}^{l} & = \widehat{\mathbf{X}}_{P,k}^{l} + \frac{\sum_{i=1}^N \mathbf{X}^{l}_i}{N}, 
\end{align}
where $l=2,...,L$ is the block layer index and $k=1, 2, ..., N_p$ is the prompt token index.  $N_p, N$ denotes the number of learnable prompt tokens and embedded patches, respectively. The global spatial prompt tokens $\widehat{\mathbf{X}}_{SP,k}^{l} \in \mathbb{R}^{N_p\times D}$ is further utilized to construct the input visual prompt tokens for the next block. It is noteworthy that this averaging operation helps the model aggregate prompt tokens with explicit global features and does not introduce any additional trainable parameters during transfer learning. Furthermore, this average operator can also be replaced with better methods, such as clustering and we provide experiments in Section~\ref{sec:kmeans}.

\begin{table*}[t]
	\renewcommand\tabcolsep{6.0pt}
	\centering
        \caption{{\bf Quantitative results of visual prompt tuning of SSL pre-trained vision transformers on FGVC datasets.} Total Params denotes the total number of parameters for the backbone encoder ViT-B, prompt tokens, and the task heads.}
   \label{tab: exp_sota_fgvc}
	\scalebox{0.8}{
		\begin{tabular}{lccccccc}
			\toprule
			Method & Total Params & CUB & Flowers & Cars & Dogs & NABirds & Average \\ 		
			\midrule
                \multicolumn{8}{l}{\it MAE Pre-train ViT-B/16:} \\
			VPT-Shallow~\citep{jia2022vpt} & 1.02x & 42.15 & 69.15 & 43.38 & 77.07 & 57.43 & 57.84 \\
                VPT-Deep~\citep{jia2022vpt} & 1.02x & 68.33 & 80.05 & 67.67 & 78.83 & 65.22 & 72.02  \\
                GaPT~\citep{yoo2023improving} & 1.02x & 70.56 & 78.55 & 71.70 & 78.90 & 67.26 & 73.39 \\
                \method (ours) & 1.08x & \bf 73.86 & \bf 82.32 & \bf 74.75 & \bf 82.05 & \bf 71.73 & \bf 76.94 \\
                \hline
                \multicolumn{6}{l}{\it MoCo v3 Pre-train ViT-B/16:} \\
                VPT-Shallow~\citep{jia2022vpt} & 1.02x & 79.05 & 90.47 & 71.91 & 81.97 & 72.92 & 79.26 \\
                VPT-Deep~\citep{jia2022vpt} & 1.02x & 82.67 & 94.41 & 79.18 & 83.33 & 75.99 & 83.12  \\
                GaPT~\citep{yoo2023improving} & 1.02x & 82.86 & 93.71 & 79.02 & 83.37 & 76.02 & 83.00 \\
                \method (ours) & 1.08x & \bf  84.29 & \bf 95.06 & \bf 80.12 & \bf 84.25 & \bf 77.16 & \bf 84.18 \\  
			\bottomrule
			\end{tabular}}
  \vspace{-0.5em}
\end{table*}

\subsection{Long-term Prompt Coding}\label{sec:lpc}
To explicitly learn from prompt sources across long-range blocks, we introduce a novel long-term prompt coding mechanism to mitigate the forgetting of previously learned prompts corresponding to objects in the image. 
Specifically, we leverage a learnable temporal coding layer consisting of long short-term memory (LSTM)~\citep{hochreiter1997long} with dimension size of $D$ to avoid forgetting the spatial prompt tokens $\widehat{\mathbf{X}}_{SP}^{l}$ introduced in the previous blocks while injecting new learnable prompt tokens $\mathbf{P}^l$.

For LSTM, given an input at current step $t$ and the hidden state $h_{t-1}$ and context state $c_{t-1}$ at time step $t-1$, the final output with hidden state $h_{t}$ and context state $c_{t}$ is defined as
\begin{equation}
\begin{aligned}
h_t,c_t &= \mbox{LSTM}(h_{t-1},c_{t-1},x_t)
\end{aligned}
\end{equation}

With the inserted prompt tokens $\mathbf{P}^l\in\mathbb{R}^{N_p\times D}$ and $\widehat{\mathbf{X}}_{SP}^{l}\in\mathbb{R}^{N_p\times D}$, we take the inserted prompts $\mathbf{P}^{l}$ as input and $\widehat{\mathbf{X}}_{SP}^{l}$ as hidden states, and the output context embeddings $\mathbf{C}^{l-1}\in\mathbb{R}^{N_p\times D}$ at block $l-1$ are fed into the layer as cell states.
Finally, the output prompt tokens $\mathbf{X}_P^{l}$ are used as the new prompts  for this long-term prompt coding, which is formulated as
\begin{equation}
\begin{aligned}
\mathbf{X}_{P}^{l},\mathbf{C}^{l} &= \mbox{LSTM}(\widehat{\mathbf{X}}_{SP}^{l},\mathbf{C}^{l-1},\mathbf{P}^{l})
\end{aligned}
\end{equation}
where $\mathbf{X}_{P}^{l},\mathbf{C}^{l}\in\mathbb{R}^{N_p\times D}$ denote the updated prompts and context embeddings, respectively. 
$D$ denotes the dimension of embeddings, and $\mathbf{P}^{l}$ is new learnable parameters inserted in the block.
$\mbox{LSTM}[\cdot]$ is the LSTM layer operator. 
After spatial prompt coding in block $l$, the LSTM layer takes the inserted prompts $\mathbf{P}^{l}$ from block $l$ as input sequences and class-ware spatial prompts $\widehat{\mathbf{X}}_{SP}^{l}$ from block $l$ as hidden states, and uses the previous output context $\mathbf{C}^{l-1}$ at block $l-1$ as cell states to generate the final prompt $\mathbf{X}_{P}^{l}$ as input prompt tokens to block $l+1$.

Note that the LSTM layer for long-term temporal coding is used starting from the first block and ending before the last block. 
With weights-specific LSTM layers for each block, we do not see obvious performance gains but introduce $(L-1)\times$ trainable parameters.
To balance the performance and total tunable parameters, we apply one weights-shared LSTM layer for efficient visual prompt tuning. 

\subsection{Overall Framework}

Overall, \method is optimized in an end-to-end manner with global spatial coding and long-term temporal coding together. Specifically, the attention block will first take the class token, prompt tokens and patch tokens from the last block to generate intermediate output tokens. Next, global spatial prompt coding is used to weave in patch tokens as spatial coding elements into output prompt tokens. Then, long-term prompt coding is employed to actively mitigate the forgetting of earned prompt tokens from previous self-attention blocks. Finally, the enhanced prompt tokens by two modules is sent to the next attention block. Besides, $\mathbf{P}^{1}$ is used as $\mathbf{X}_{P}^{1}$ for the first block, and the class token learned from the transformer is used for the downstream classification. The full algorithm is in Appendix~\ref{sec:appendix_algo}.

\begin{table*}[t]
	\renewcommand\tabcolsep{6.0pt}
	\centering
        \caption{{\bf Quantitative results of visual prompt tuning of SSL pre-trained vision transformers on VTAB-1K benchmarks.} Total Params denotes the total number of parameters for the backbone encoder ViT-B, prompt tokens, and the task heads.}
   \label{tab: exp_sota_vtab}
	\scalebox{0.8}{
		\begin{tabular}{lcccccc}
			\toprule
			Method & Total Params & Natural (7) & Specialized (4) & Structured (8) & Average \\ 		
			\midrule
                \multicolumn{6}{l}{\it MAE Pre-train ViT-B/16:} \\
			Linear & 1.01x & 18.87 & 53.72 & 23.70 & 28.24 \\
                Adapter & 1.17x & \bf 54.90 & 75.19 & 38.98 & 52.47 \\
                VPT-Shallow~\citep{jia2022vpt} & 1.01x & 39.96 & 69.65 & 27.50 & 40.96 \\
                VPT-Deep~\citep{jia2022vpt} & 1.01x & 36.02 & 60.61 & 26.57 & 37.22 \\
                GaPT~\citep{yoo2023improving} & 1.01x & 47.61 & 76.86 & 36.80 & 49.22 \\
                \method (ours) & 1.05x &  52.36 & \bf 80.75 & \bf 41.72 & \bf 53.86 \\ \hline
                \multicolumn{6}{l}{\it MoCo v3 Pre-train ViT-B/16:} \\
                Linear & 1.01x & 67.46 & 81.08 & 30.33 & 54.69 \\
                Adapter & 1.22x & 74.19 & 82.66 & 47.69 & 64.82 \\
                VPT-Shallow~\citep{jia2022vpt} & 1.01x &                  67.34 & 82.26 & 37.55 & 57.94 \\
                VPT-Deep~\citep{jia2022vpt} & 1.01x & 70.27 & 83.04 & 42.38 & 61.22 \\
                GaPT~\citep{yoo2023improving} & 1.01x & 74.84 & 83.38 & 49.10 & 65.80 \\
                 \method (ours) & 1.05x &  \bf 77.19 & \bf 85.69 & \bf 52.82 & \bf 68.72 \\
			\bottomrule
			\end{tabular}}
  \vspace{-1.0em}
\end{table*}

\begin{table}[t]
	\renewcommand\tabcolsep{4.0pt}
	\centering
        \caption{{\bf Quantitative results of visual prompt tuning of SSL pre-trained vision transformers on ADE-20K for semantic segmentation.} SS and MS denote single-scale and multi-scale, respectively.}
   \label{tab: exp_seg}
	\scalebox{0.68}{
		\begin{tabular}{lcccccc}
			\toprule
			\multirow{2}{*}{Method}  & \multicolumn{2}{c}{MAE} & \multicolumn{2}{c}{MoCo v3} \\
               & mIoU (SS) &   mIoU (MS) & mIoU (SS) &   mIoU (MS) \\	
			\midrule
   VPT-Shallow~\citep{jia2022vpt} & 34.20 & 35.23 & 34.55 & 36.18 \\
   VPT-Deep~\citep{jia2022vpt} & 37.76 & 38.80 & 35.50 & 37.15 \\
   GaPT~\citep{yoo2023improving} & 38.44 & 39.81 & 36.81 & 38.55 \\
   LSPT (ours) & \bf 39.72 & \bf 41.51 & \bf 37.92 & \bf 39.73 \\
			\bottomrule
			\end{tabular}}
  \vspace{-0.5em}
\end{table}

\begin{table*}[t]
	\renewcommand\tabcolsep{6.0pt}
	\centering
        \caption{{\bf Ablation studies on Global Spatial Prompt Coding (GSPC) and Long-term Prompt Coding (LPC).}}
   \label{tab: exp_ablation}
	\scalebox{0.8}{
		\begin{tabular}{cccccccccc}
			\toprule
			GSPC & LPC & CUB & Flowers & Cars & Dogs & NABirds & Natural (7) & Specialized (4) & Structured (8)\\ 
			\midrule
                \xmark & \xmark & 68.33 & 80.05 & 67.67 & 78.83 & 65.22 & 67.34 & 82.26 & 37.55 \\
                \cmark & \xmark & 76.28 & 85.23 & 72.16 & 80.51 & 70.63 & 72.36 & 83.56 & 45.15 \\
                \xmark & \cmark & 78.12 & 89.17 & 75.38 & 82.75 & 73.58 & 75.18 & 84.28 & 48.39 \\
			\cmark & \cmark & \bf  84.29 & \bf 95.06 & \bf 80.12 & \bf 84.25 & \bf 77.16 & \bf 77.19 & \bf 85.69 & \bf 52.82 \\  
			\bottomrule
			\end{tabular}}
  \vspace{-0.5em}
\end{table*}

\section{Experiments}
In this section, we will introduce the experiments conducted by us to answer the following research questions:

\paragraph{Q1.} How well does our \method perform on transfer learning benchmarks compared to the previous visual prompting baselines?

\paragraph{Q2.} To what extent does the global spatial prompt coding and long-term prompt coding contribute to the final performance?

\paragraph{Q3.} Does the global spatial prompt coding and long-term prompt coding help address the spatial and temporal forgetting problem?

\subsection{Experimental Setup}
We first introduce the dataset, evaluation metrics we used and our implementation for the experiments.

\paragraph{Datasets.} Our experiments are conducted on two widely used classification datasets, FGVC and VTAB-1K.

FGVC benchmark includes 5 fine-grained classification tasks: CUB-200-2011~\citep{wah2011cub}, Oxford Flowers~\citep{Nilsback2008flowers}, Stanford Cars~\citep{Gebru2017cars}, Stanford Dogs~\citep{Khosla2011dogs}, and NABirds~\citep{van2015birds}.
Following the prior work~\citep{jia2022vpt,yoo2023improving}, we use the same split for training and validation.

VTAB-1K~\citep{zhai2019largescale} dataset consists of 19 diverse visual classification tasks, and is composed of three subgroups: Natural with natural images obtained from standard cameras, Specialized with images captured using specialized equipments (medical and satellite imagery), and Structured which requires geometric understanding such as object counting. Each task contains 1000 training examples, and we use the same split in~\citep{jia2022vpt,yoo2023improving} to run the final training and evaluation.

\begin{table}[ht]
	\renewcommand\tabcolsep{6.0pt}
	\centering
        \caption{{\bf Ablation studies on Long-term Prompt Coding (LPC) using LSTM vs Transformer.}}
   \label{tab: exp_transformer}
	\scalebox{0.8}{
		\begin{tabular}{lcccccc}
			\toprule
			LPC      & CUB            & Flowers        & Cars           & Dogs           & NABirds        & AVG   \\	
			\midrule
                Transformer & 72.52 & 81.26   & 73.58 & 81.23 & 70.86   & 75.89 \\
LSTM        & \bf 73.86 & \bf 82.32   & \bf 74.75 & \bf 82.05 & \bf 71.73   & \bf 76.94 \\
			\bottomrule
			\end{tabular}}
  \vspace{-1.0em}
\end{table}

\begin{table}[t]
	\renewcommand\tabcolsep{6.0pt}
	\centering
        \caption{{\bf Ablation studies on Global Spatial Prompt Coding (GSPC) using $k$-means.}}
   \label{tab: exp_kmeans}
	\scalebox{0.78}{
		\begin{tabular}{lcccccc}
			\toprule
			Spatial Prompt       & CUB            & Flowers        & Cars           & Dogs           & NABirds        & AVG   \\	
			\midrule
Average        & 73.86          & 82.32          & 74.75          & 82.05          & 71.73          & 76.94          \\
$k$-means        & \bf 74.32 & \bf 82.56 & \bf 74.87 & \bf 82.23 & \bf 71.86 & \bf 77.17 \\
			\bottomrule
			\end{tabular}}
  \vspace{-0.5em}
\end{table}

\paragraph{Evaluation Metrics.}
For FGVC and VTAB-1K benchmarks, we report the individual and the average accuracy on the datasets.
For individual results on each benchmark, we compute the average accuracy score on the test set within three runs using different seeds.
For VTAB-1K, we report the average accuracy score on three subgroups and the overall dataset.

\paragraph{Implementation.}
We apply ViT-B/16 as the backbone architecture in all experiments.
For self-supervised vision transformers, we use MAE~\citep{he2021masked} and MoCo v3~\citep{chen2021mocov3} pre-trained on ImageNet-1K~\citep{imagenet_cvpr09}.
We follow the same pre-trained model parameters as the prior work~\citep{jia2022vpt,yoo2023improving}.

\begin{figure*}[t]
\centering
\includegraphics[width=0.95\linewidth]{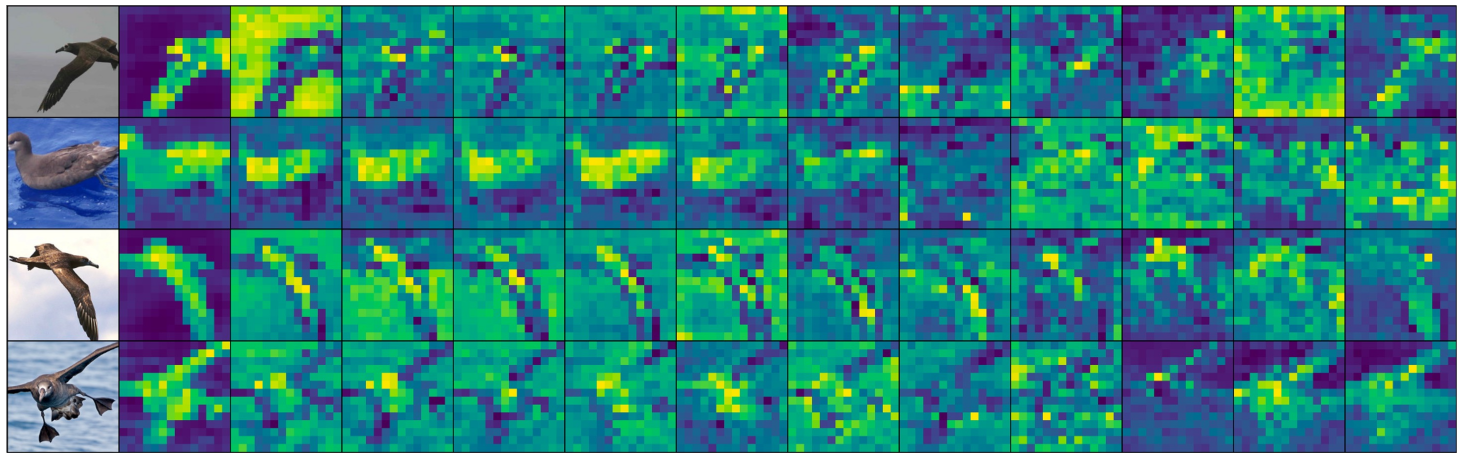}
\vspace{-0.5em}
\caption{{\bf Qualitative visualization of long-term prompt forgetting in state-of-the-art visual prompt tuning method~\citep{yoo2023improving}.} From left to right: layer 1 to layer 12.}
\label{fig: vis_temporal_forget}
\vspace{-0.5em}
\end{figure*}

\begin{figure*}[t]
\centering
\includegraphics[width=0.95\linewidth]{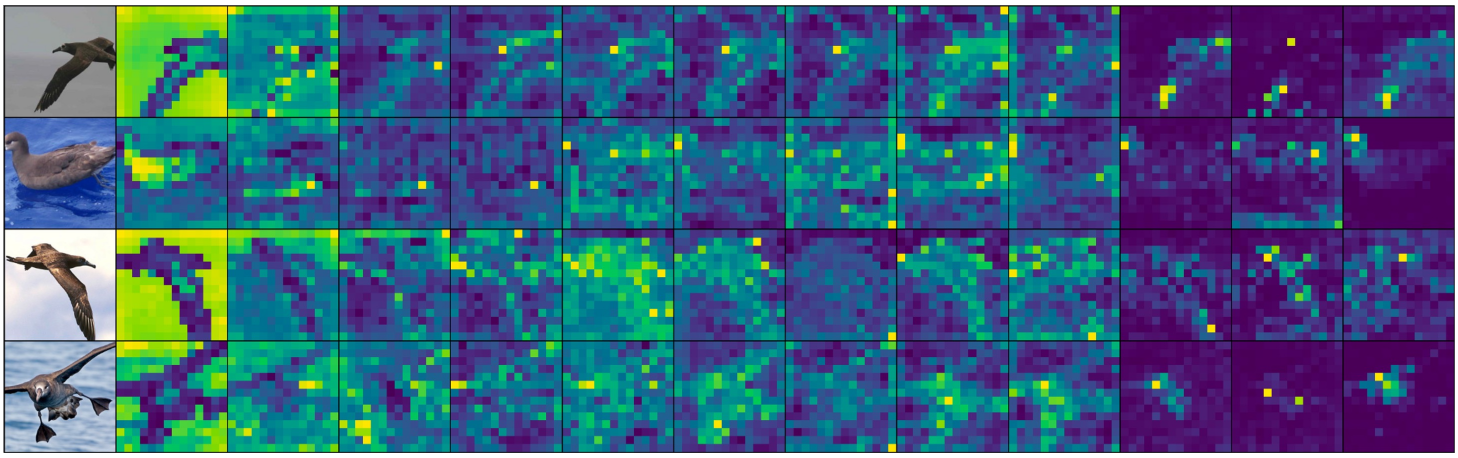}
\vspace{-0.5em}
\caption{{\bf Qualitative visualization of spatial attention forgetting in state-of-the-art visual prompt tuning method~\citep{yoo2023improving}.} From left to right: layer 1 to layer 12.}
\label{fig: vis_spatial_forget}
\end{figure*}

\subsection{Comparison to Prior Work}\label{sec:exp_sota}
To answer \textbf{Q1} and demonstrate the effectiveness of the proposed \method, we comprehensively compare it to previous finetuning and visual prompt tuning 
baselines:
1) Linear: a vanilla baseline that used a linear layer as the classification head;
2) Adapter~\citep{pfeiffer2020adapterhub}: a strong baseline that inserted learnable MLP modules with residual connection across transformer blocks; 
3) VPT~\citep{jia2022vpt}: the first baseline that prepended learnable prompt tokens to the input sequences as task-specific instructions for prompt tuning. Both variants (VPT-Shallow and VPT-Deep) are listed for comparison;
4) GaPT~\citep{yoo2023improving}: a recent and strong baseline with gated prompts and adaptive attention.

For the FGVC datasets, we report the quantitative comparison results in Table~\ref{tab: exp_sota_fgvc}.
As can be seen, we achieve the best results regarding all metrics for five fine-grained classification tasks compared to previous visual prompt tuning approaches using MAE and MoCo v3 pre-trained weights.
In particular, the proposed \method superiorly outperforms GaPT~\citep{yoo2023improving}, the current state-of-the-art visual prompt tuning baseline, by 3.30@CUB, 3.77@Flowers, 3.05@Cars, 3.15@Dogs, and 4.47@NABirds, when evaluated on MAE pre-trained weights.
Furthermore, we achieve significant performance gains compared to VPT~\citep{jia2022vpt}, the first visual prompt tuning baseline, which indicates the importance of explicitly mitigating the forgetting of prompt tokens learned from history blocks for effective prompt tuning.
These significant improvements demonstrate the superiority of our approach in visual prompt tuning.

In addition, significant gains in VTAB-1K benchmarks can be observed in Table~\ref{tab: exp_sota_vtab}.
Compared to GaPT~\citep{yoo2023improving}, the current state-of-the-art visual prompt tuning baseline, we achieve the results gains of 2.35@Natural, 2.31@Specialized, and 3.72@Structured in terms of MoCo v3 pre-trained weights.
Moreover, when evaluated on the challenging Structured datasets using MAE pre-trained weights, the proposed method still outperforms GaPT~\citep{yoo2023improving} by 4.92.
We also achieve highly better results against Linear and VPT~\citep{jia2022vpt}.
These results demonstrate the effectiveness of our approach in learning long-term and global prompts for visual prompt tuning on downstream classification.

Pre-trained vision transformers are designed to be applied on a variety of downstream applications. Besides image classification, we would like to see whether our proposed LSPT can also benefit other tasks, such as semantic segmentation.
To this end, we follow previous work~\citep{jia2022vpt,yoo2023improving} and train SETR-PUP~\citep{zheng2021rethinking} model as the segmentation transformer framework on ADE20K dataset~\citep{zhou2017scene,Zhou2018SemanticUO}.
Table~\ref{tab: exp_seg} reports the comparison results with previous visual prompt tuning approaches~\citep{jia2022vpt,yoo2023improving} using MAE and MoCo v3 pre-trained ViT-B/16 weights.
As can be seen, our \method achieves the best performance in terms of all metrics for two different pre-trained models.
These significant improvements demonstrate the superiority of our framework in visual prompting on semantic segmentation.

\subsection{Ablation Studies}

In this section, we try to answer \textbf{Q2} and performed ablation studies to demonstrate the benefit of introducing Global Spatial Prompt Coding (GSPC) and Long-term Prompt Coding (LPC) modules. 
We ablate the necessity of each module and report the quantitative results on all downstream datasets using MoCo v3 pre-trained weights in Table~\ref{tab: exp_ablation}.
As can be observed, adding GSPC to the vanilla baseline highly increases the results by 9.79@CUB, 9.12@Flowers, 7.71@Cars, 3.92@Dogs, 8.36@NABirds, 7.84@Natural, 2.02@Specialized, and 10.84@Structured, which validates the benefit of long-term prompt coding in learning long-range blocks as prompt sources for visual prompt tuning.
Similarly, introducing only GSPC in the baseline increases the prompt tuning performance regarding all metrics.
More importantly, incorporating both LPC and GSPC into the baseline significantly raises the performance by 15.96@CUB, 15.01@Flowers, 12.45@Cars, 5.42@Dogs, 11.94@NABirds, 9.85@Natural, 3.43@Specialized, and 15.27@Structured.
These improving results validate the parallel importance of mitigating the forgetting of prompt tokens learned from history blocks temporally and image patch tokens spatially for visual prompt tuning.

\subsection{Ablation on LSTM vs Transformer in Long-term Prompt Coding}
In order to further demonstrate the effectiveness of using LSTM for long-term modeling, we ablated the Long-term Prompt Coding module by using a transformer layer to aggregate previous and new prompt tokens.
Table~\ref{tab: exp_transformer} reports the comparison results on 5 fine-grained visual classification datasets.
As can be observed, replacing LSTM with a transformer layer deteriorates the results in terms of all benchmarks. 
This might be because the auto-aggressive transformer layer applied among tokens increases their similarity by a weighted sum of token values, leading to losing discrepancy among latent toke embeddings during training.  
In contrast, the choice of LSTM for long-term modeling helps introduce a forget gate to enlarge the variance for prompt tokens.

\subsection{Ablation on Global Spatial Prompt Coding}\label{sec:kmeans}

Adding the average of aggregated patch tokens from attention blocks is beneficial for accumulating spatial and positional information from self-supervised ViT attention weights.
To explore such effects more comprehensively, we replaced the average operator with a $k$-means clustering on input patch tokens and report the quantitative results in Table~\ref{tab: exp_kmeans}.
Specifically, $k$ is set to match the number of visual prompts, and the centroids of clusters are added to prompt tokens as Global Spatial Prompt Coding (GSPC). Our \method with $k$-means clustering achieves better results, which demonstrates the importance of spatial prompt coding in visual prompt tuning on specific downstream tasks.
This advanced mechanism allows for a more detailed understanding of spatial relationships within the data, which is particularly beneficial for tasks that require fine-grained differentiation between classes.
However, the additional computational cost will be taken on the $k$-means clustering for training time. 
Therefore, more exploration space on this spatial prompt coding will leave for future work.

\subsection{Visualization of Attention Maps}
We introduce the temporal and spatial forgetting problem in the state-of-the-art VPT method and answer \textbf{Q3} by visualizing the similarity between prompt tokens and the patch tokens as well as the attention maps in the blocks. With our Long-term Prompt Coding and Global Spatial Prompt Coding, we expect to see a clear awareness of the target object from the visualization even in the very posterior blocks.

\noindent\textbf{Long-term and Spatial Prompt Forgetting in State-of-the-art VPT.}
In order to validate long-term and spatial prompt forgetting in GaPT~\citep{yoo2023improving}, the state-of-the-art visual prompt tuning approach, we compute the cosine similarity between learnable prompt tokens and patch embeddings, and average the results along the number of prompt tokens. 
The qualitative visualization maps across 12 transformer layers are showcased in Figure~\ref{fig: vis_temporal_forget}, which corresponds to the temporal forgetting across blocks.
As can be seen in the first column, prompt tokens in the first layer attend to learn features corresponding to objects in the image.
However, prompt tokens in the last few layers fail to capture objects in the image as they do not explicitly mitigate the forgetting of prompt tokens learned from history blocks. 
Meanwhile, we visualize the spatial attention maps on the averaged head from 12 self-attention layers in Figure~\ref{fig: vis_spatial_forget}, which corresponds to the spatial forgetting on each transformer block.
We can also observe that the spatial attention maps become worse when it comes to the last few layers since there is no direct connection between the learnable prompt tokens and local patch embeddings.

\section{Conclusion}

In this work, we navigated the multifaceted landscape of visual representation learning, pinpointing the existing gaps and challenges that have persisted, particularly in the realm of Visual Prompt Tuning (VPT). 
Our observations laid bare the limitations of extant methods, which, despite their efficacy, struggled to exploit the rich information embedded in long-range blocks and patch tokens of self-supervised Vision Transformers.
Rising to address this need, we introduced \methodfull (\method) with the Long-Term Prompt Coding and the Global Spatial Prompt Coding which firmly establishes \method's capabilities in both retaining pivotal parameters from earlier blocks and continually aggregating class-centric features.
The comprehensive evaluations on diverse benchmarks underscore \method's unmatched prowess, as it consistently outperformed prevailing baselines, setting new standards for visual prompt tuning performance.

\newpage

\section*{Impact Statement}

LSPT is a preliminary work to address the temporal forgetting problem and the spatial forgetting problem within the visual prompt tuning techniques. While we compared several designs for long-term prompt coding and global spatial prompt coding in ablation studies, most choices are rather effective yet straight-forward. However, we believe there is still much room to obtain a better balance between cost and effectiveness, which we leave for future exploration.

Since prompt tuning is a widely used concept shared across modalities, potential future directions include exploring language-guided VPT for cross-modal understanding where both images and languages are involved. This may lead to a more holistic view of LSPT, extend its potential applications and areas for wider research community.

\bibliography{reference}
\bibliographystyle{icml2024}

\newpage
\appendix
\onecolumn

\appendix

\counterwithin{figure}{section}
\counterwithin{table}{section}
\counterwithin{algorithm}{section}

\section*{Appendix}

In this appendix, we further provide the following materials:
\begin{itemize}
    \item additional experiments on medical images and supervised ImageNet-21k in Section~\ref{sec:appendix_exp},
    \item the pseudo-algorithm framework of our \method in Section~\ref{sec:appendix_algo}, 
    \item additional analyses on the number of LSTM \& GRU layers, training \& inference costs, and visualization of attention maps in Section~\ref{sec:appendix_ablation}.
\end{itemize}

\section{Additional Experiments}\label{sec:appendix_exp}

In order to further demonstrate the effectiveness of the proposed \method in visual prompt tuning, we conduct experiments on  
medical visual adaptation benchmark and image classification using supervised ImageNet-21K ViT-B/16 models on VTAB-1K datasets.

\begin{table*}[t]
	\renewcommand\tabcolsep{6.0pt}
	\centering
        \caption{{\bf Quantitative results of visual prompt tuning of SSL pre-trained vision transformers on nine medical images benchmarks.} Total Params denotes the total number of parameters for the backbone encoder ViT-B, prompt tokens, and the task heads.}
   \label{tab: exp_sota_med}
	\scalebox{0.85}{
		\begin{tabular}{lc|ccc|ccc|cc|c}
			\toprule
			\multirow{2}{*}{Method}     & Total & HyperKvasir & MESAD & Kvasir & AMLC & LHNCBC & MLLBone & APTOS & EyePACS & ISIC\\
            & Params & Polyp & Prostatectomy & Polyp & Cell & Cell & Cell & Eye & Eye & Skin  \\
			\midrule
                \multicolumn{10}{l}{\it MAE Pre-train ViT-B/16:} \\
                VPT-Shallow & 1.02X  & 52.32 & 33.16 & 59.06 & 27.29 & 34.25 & 23.29 & 49.76 & 42.16 & 42.53 \\
                VPT-Deep    & 1.02X  & 56.13 & 37.53 & 63.17 & 30.51 & 39.32 & 29.86 & 53.62 & 45.21 & 46.79 \\
                GaPT        & 1.02X  & 59.32 & 41.67 & 65.25 & 31.68 & 41.53 & 31.96 & 56.13 & 49.32 & 48.53 \\
                LSPT (ours)        & 1.05X  & \bf 61.67 & \bf 43.26 & \bf 68.17 & \bf 33.25 & \bf 43.78 & \bf 33.25 & \bf 58.92 & \bf 51.28 & \bf 50.17 \\ \hline
                \multicolumn{10}{l}{\it MoCo v3 Pre-train ViT-B/16:} \\
                VPT-Shallow & 1.02X  & 55.82 & 35.78 & 61.23 & 29.15 & 36.82 & 26.57 & 51.67 & 44.26 & 45.06 \\
                VPT-Deep    & 1.02X  & 58.27 & 39.05 & 65.21 & 32.81 & 40.05 & 31.02 & 54.78 & 46.79 & 48.65 \\
                GaPT         & 1.02X  & 61.24 & 42.31 & 66.02 & 34.57 & 42.36 & 33.52 & 57.09 & 50.23 & 49.72 \\
                LSPT (ours)        & 1.05X  & \bf 63.57 & \bf 44.52 & \bf 68.73 & \bf 35.72 & \bf 45.21 & \bf 35.71 & \bf 59.33 & \bf 51.76 & \bf 51.26 \\
			\bottomrule
			\end{tabular}}
  \vspace{-0.5em}
\end{table*}

\subsection{Medical Visual Adaptation}

Medical visual adaptation is a challenging problem that uses medical color images as input to transfer our general pre-trained visual transformers for medical downstream classification tasks.
We follow previous work~\citep{jia2022vpt,yoo2023improving}, and report the comparison results with previous visual prompt tuning methods~\citep{jia2022vpt,yoo2023improving} using MAE and MoCo v3 pre-trained ViT-B/16 models on nine medical image datasets from diverse organs, including polyp (HyperKvasir~\cite{borgli2020HyperKvasir}, MESAD Prostatectomy~\citep{Bawa2021ESAD}, Kvasir~\citep{Kvasirv2}), cell (AMLC~\cite{matek2019human}, LHNCBC~\cite{Lhncbc}, MLLBone~\cite{Matek2021highly}), eye (APTOS~\cite{aptos}, EyePACS~\cite{eyepacs}), and skin (ISIC~\cite{isic}).
As can be observed in Table~\ref{tab: exp_sota_med}, our \method achieves the best performance regarding all datasets for two different pre-trained models.
These significant improvements demonstrate the generalizability of our framework in visual prompting on medical images.

\subsection{Supervised ImageNet-21k.}
To validate the generalizability of the proposed \method on visual prompt tuning using supervised weights, we comprehensively compare it with current visual prompt tuning approaches~\citep{das2023learning,wang2023adapting,jie2023revisiting} in supervised settings of ImageNet-21K  ViT-B/16 models on VTAB-1K benchmarks~\citep{zhai2019largescale}.
The comparison quantitative results are shown in Table~\ref{tab: exp_sup}.
Compared to previous methods, we achieve the best results regarding all various benchmarks, including natural, specialized, and structured. 
In particular, the proposed \method outperforms Bi-AdaptFormer~\citep{jie2023revisiting}, the state-of-the-art method by minimizing prompts quantization errors, by 3.15@Natural, 2.17@Specialized, 3.82@Structured. 
We also achieve highly better results against EXPRES~\citep{das2023learning} that tried to learn residual tokens for the output of various computations.
These results validate the effectiveness of our approach in visual prompt tuning on supervised weights.

\begin{table*}[ht]
	\renewcommand\tabcolsep{6.0pt}
	\centering
        \caption{{\bf Quantitative results of visual prompt tuning of supervised ImageNet-21K ViT-B/16 weights on VTAB-1K benchmarks.} Numbers in ($\cdot$) denote the number of downstream datasets.}
   \label{tab: exp_sup}
	\scalebox{0.9}{
		\begin{tabular}{lcccc}
			\toprule
			Method & Natural (7) & Specialized (4) & Structured (8) & Average \\ 		
			\midrule
			VPT-Shallow~\citep{jia2022vpt} & 76.81 & 79.66 & 46.98 & 64.85 \\
                VPT-Deep~\citep{jia2022vpt} & 78.48 & 82.43 & 54.98 & 69.42 \\
                EXPRES~\citep{das2023learning} & 79.70 & 84.00 & 55.00 & 70.21 \\
                SNF~\citep{wang2023adapting} & 83.79 & 86.13 & 59.61 & 74.10 \\
                Bi-AdaptFormer~\citep{jie2023revisiting} & 82.11 & 86.40 & 62.43 & 74.73 \\
                LSPT (ours) & \bf 85.26 & \bf 88.57 & \bf 66.25 & \bf 77.95 \\
			\bottomrule
			\end{tabular}}
\end{table*}

\begin{algorithm}[t]
\caption{{\bf Long-term Spatial Prompt Tuning}}\label{alg:lspt}
\begin{algorithmic}
\Require input patch tokens for $l$th self-attention block $\mathbf{X}^{l-1}\in\mathbb{R}^{N\times D}$, classification tokens $\mathbf{x}^{l-1}_C\in\mathbb{R}^{1\times D}$, inserted prompt tokens $\mathbf{P}^{l}\in\mathbb{R}^{N_p\times D}$, $l$th ViT blocks $\mbox{AttnBlock}^l(\cdot)$, a single LSTM layer $\mbox{LSTM}(\cdot)$, number of patch tokens $N$, number of prompt tokens $N_p$, number of blocks $L$.
\For{$l=1,2,...,L$} 
    \State $\mathbf{x}_C^{l}, \hat{\mathbf{X}}_P^{l}, \mathbf{X}^{l} \gets \mbox{AttnBlock}^l(\mathbf{x}_C^{l-1}, \mathbf{X}_P^{l-1}, \mathbf{X}^{l-1})$ 
    \State $\hat{\mathbf{X}}_{SP,k}^{l} \gets \hat{\mathbf{X}}_{P,k}^{l} + \dfrac{\sum_{i=1}^N\mathbf{X}_i^{l}}{N}$  \Comment{Global Spatial Prompt Coding} 
    \State $\mathbf{X}_{P}^{l}, \mathbf{C}^{l} \gets \mbox{LSTM}(\hat{\mathbf{X}}_{SP}^{l},\mathbf{C}^{l-1}, \mathbf{P}^l$) \Comment{Long-term Prompt Coding} 
\EndFor
\State \Return $\mathbf{X}_{P}^{l}$
\end{algorithmic}
\end{algorithm}
\vspace{-1.0em}

\section{Pseudo Algorithm for \method}\label{sec:appendix_algo}

To enhance the replicability of our method, we report the pseudo algorithm of our \method in Algorithm~\ref{alg:lspt}. 
Specifically, we are given input patch tokens for $l$th self-attention block $\mathbf{X}^{l-1}\in\mathbb{R}^{N\times D}$, classification tokens $\mathbf{x}^{l-1}_C\in\mathbb{R}^{1\times D}$, inserted prompt tokens $\mathbf{P}^{l}\in\mathbb{R}^{N_p\times D}$, $l$th ViT blocks $\mbox{AttnBlock}^l(\cdot)$, a single LSTM layer $\mbox{LSTM}(\cdot)$, where $N, N_p, L$ denote the number of patch tokens, number of prompt tokens, and number of blocks, respectively.
For each transformer block $l$, we first apply ViT self-attention blocks $\mbox{AttnBlock}^l(\cdot)$ to aggregate input classification tokens $\mathbf{x}_C^{l-1}$, prompt tokens $\mathbf{X}_P^{l-1}$, and patch tokens $\mathbf{X}^{l-1}$ for generating the output classification tokens $\mathbf{x}_C^{l}$, prompt tokens $\hat{\mathbf{X}}_P^{l}$, and and patch tokens $\mathbf{X}^{l}$.

After aggregation, we simply apply two proposed modules including the Global Spatial Prompt Coding (GSPC), and Long-term Prompt Coding (LPC) on the output tokens $\hat{\mathbf{X}}_P^{l},\mathbf{X}^{l}$, newly inserted prompt tokens $\mathbf{P}^{l}$, and context tokens $\mathbf{C}^{l-1}$ from last layer.
The Global Spatial Prompt Coding (GSPC) module adds the average embeddings of patch tokens $\mathbf{X}^{l}\in\mathbb{R}^{N\times D}$ from the block to the output prompts $\widehat{\mathbf{X}}_P^{l}\in\mathbb{R}^{N_p\times D}$ to generate global spatial prompts $\widehat{\mathbf{X}}_{SP}^{l}$.
With the inserted prompt tokens $\mathbf{P}^l\in\mathbb{R}^{N_p\times D}$ and $\widehat{\mathbf{X}}_{SP}^{l}\in\mathbb{R}^{N_p\times D}$, the Long-term Prompt Coding (LPC) module with parallel importance takes the inserted prompts $\mathbf{P}^{l}$ as input and $\widehat{\mathbf{X}}_{SP}^{l}$ as hidden states, and the output context embeddings $\mathbf{C}^{l-1}\in\mathbb{R}^{N_p\times D}$ at block $l-1$ are fed into the layer as cell states.
Finally, the output updated prompts $\mathbf{X}_P^{l}$ are used as the new prompt tokens for block $l+1$ to achieve long-term prompt coding.

\section{Additional Analysis}\label{sec:appendix_ablation}

In this section, we performed ablation studies to demonstrate the advantage of using one single LSTM layer in the Long-term Prompt Coding (LPC) module against Gated Recurrent Unit (GRU) and computational costs for training and inference.
Our ablation experiments are based on MAE pre-trained ViT-B/16 models.

\subsection{Ablation on Number of LSTM \& GRU Layers}
To validate the effectiveness of using a single LSTM layer as the Long-term Prompt Coding (LPC) module, we varied the number of LSTM layers from $\{1,2\}$ and ablated the layer using a single Gated Recurrent Unit (GRU) layer.
The comparison results are reported in Table~\ref{tab: exp_gru}.
We can observe that adding one more LSTM layer to our current \method achieves better results, which indicates the importance of LSTM in alleviating long-term forgetting problems for visual prompt tuning.
However, two LSTM layers bring more tunable parameters on computational overhead.
Meanwhile, replacing one single LSTM layer with a single GRU layer will not improve the performance although it has fewer parameters.
These results demonstrate the effectiveness of using one single LSTM layer in achieving a good trade-off between parameters and performance.

\begin{table*}[t]
	\renewcommand\tabcolsep{6.0pt}
	\centering
        \caption{{\bf Ablation studies on Long-term Prompt Coding (LPC) regarding the number of LSTM layers and GRU.}}
   \label{tab: exp_gru}
	\scalebox{0.95}{
		\begin{tabular}{lccccccc}
			\toprule
			LPC       & Params         & CUB            & Flowers        & Cars           & Dogs           & NABirds        & AVG   \\	
			\midrule
1 \# LSTM & 1.08x          & 73.86          & 82.32          & 74.75          & 82.05          & 71.73          & 76.94 \\
2 \# LSTM &  1.14x & \bf 74.57 & \bf 82.95 & \bf 75.52 & \bf 82.97 & \bf 72.45 & \bf 77.69 \\
1 \# GRU  & \bf 1.06x          & 72.95          & 81.53          & 73.91          & 81.26          & 70.97          & 76.12 \\
			\bottomrule
			\end{tabular}}
  \vspace{-0.5em}
\end{table*}

\subsection{Training \& Inference Costs}

In order to comprehensively assess the efficiency of the proposed \method, we compared it with GaPT~\cite{yoo2023improving}, the state-of-the-art visual prompt tuning method on self-supervised ViTs, on max memory usage, training time per batch and inference time per batch in Table~\ref{tab: exp_cost}.
We can observe that our \method achieves comparable computation costs in terms of all metrics, especially on inference time per batch.
More importantly, we achieve much better downstream performance regarding image classification in Table~\ref{tab: exp_sota_fgvc}\&~\ref{tab: exp_sota_vtab} and semantic segmentation in Table~\ref{tab: exp_seg}.
These computational analyses further demonstrate the efficiency of our novel framework.

\begin{table*}[!htb]
	\renewcommand\tabcolsep{6.0pt}
	\centering
        \caption{{\bf Comparsion results of training \& inference costs with the state-of-the-art visual prompt tuning approach on self-supervised ViT.}}
   \label{tab: exp_cost}
	\scalebox{0.95}{
		\begin{tabular}{lccc}
			\toprule
		\multirow{2}{*}{Method} & Max Memory & Training Time & Inference Time \\
            & Usage (GB) & per Batch (s)  & per Batch (s) \\ \midrule
GaPT~\citep{yoo2023improving} & 23.78          & 0.2406                      & 0.0871                       \\
LSPT (ours)    & 24.02         & 0.2428                      & 0.0872    \\                  
			\bottomrule
			\end{tabular}}
\end{table*}

\begin{figure*}[!htb]
\centering
\includegraphics[width=0.98\linewidth]{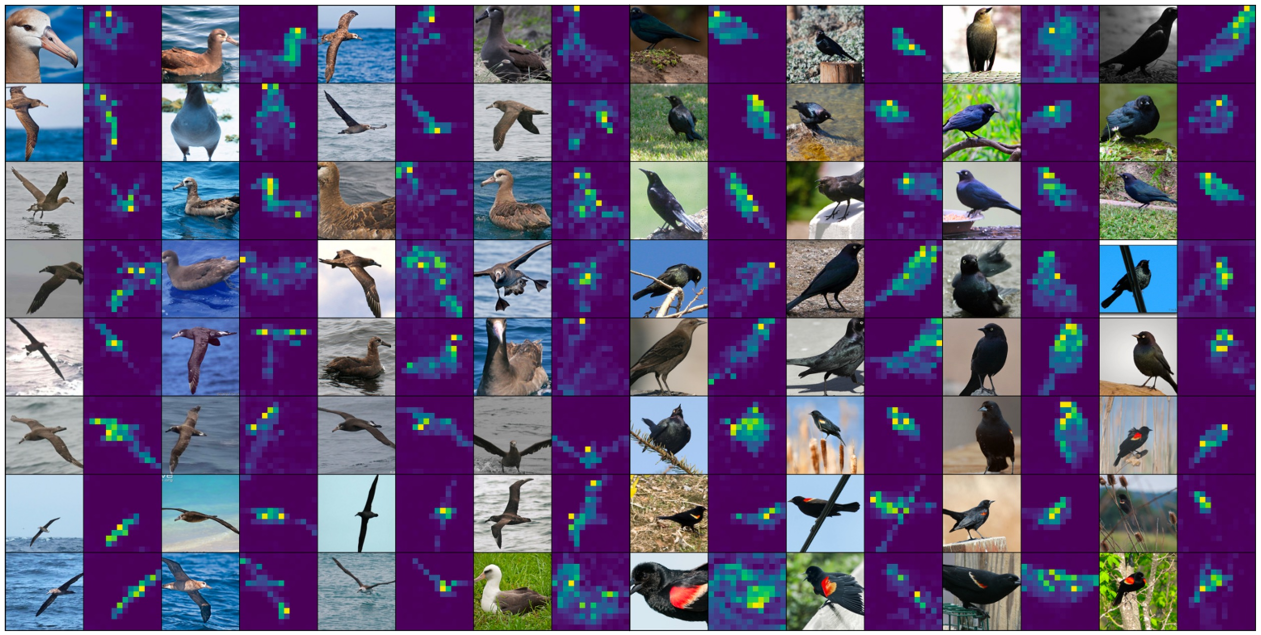}
\vspace{-0.5em}
\caption{{\bf Qualitative visualization of learned category-aware attention maps learned by the proposed \method.}}
\label{fig: exp_vis_attention}
\vspace{-1.0em}
\end{figure*}

\subsection{Visualization of Attention Maps}

\noindent\textbf{Learned Category-aware Attention Maps.}
Learning category-aware attention maps during transfer learning is essential for classifying fine-grained images.
To better evaluate the quality of learned category-aware attention maps, we visualize the learned attention maps from the last self-attention layer by averaging all heads in Figure~\ref{fig: exp_vis_attention}. 
As can be seen in the maps neighboring to the original image, attention maps extracted from our \method are discriminative to capture the shape of corresponding objects in the images.
In contrast to our discriminative maps, the spatial attention maps from GaPT~\citep{yoo2023improving} in the last column of Figure~\ref{fig: vis_spatial_forget} are blurred and coarse, where the mixture of objects and background patches still exist.
These meaningful visualization results further showcase the superiority of our \method in alleviating the forgetting of history prompt tokens obtained from self-attention transformer blocks for visual prompt tuning.


\end{document}